\documentclass[runningheads]{llncs}
\usepackage{graphicx}

\usepackage{tikz}
\usepackage{comment}
\usepackage{amsmath,amssymb} 
\usepackage{color}

\usepackage[accsupp]{axessibility}  

\usepackage{tabularx}
\usepackage{tabularx,booktabs}
\newcolumntype{C}{>{\centering\arraybackslash}X}
\usepackage{cuted}
\usepackage{capt-of}
\usepackage{adjustbox}
\usepackage{multirow}
\usepackage{pifont}

\usepackage{amssymb}
\usepackage{xcolor}

\usepackage[pagebackref=false,breaklinks=true,colorlinks,bookmarks=false,urlcolor=blue, citecolor=blue]{hyperref}

\begin{document}
\pagestyle{headings}
\mainmatter
\def\ECCVSubNumber{51}  

\title{LiteDepth: Digging into Fast and Accurate \\ Depth Estimation on Mobile Devices} 
 
\titlerunning{ECCV-22 submission ID \ECCVSubNumber} 
\authorrunning{ECCV-22 submission ID \ECCVSubNumber} 
\author{Anonymous ECCV submission}
\institute{Paper ID \ECCVSubNumber}

\author{Zhenyu Li\inst{1} \and
Zehui Chen\inst{2} \and
Jialei Xu\inst{1} \and
Xianming Liu\inst{1} \and
Junjun Jiang\inst{1}\thanks{Corresponding author (jiangjunjun@hit.edu.cn).}}
\authorrunning{Z. Li et al.}
%
\institute{Harbin Institute of Technology \and
University of Science and Technology of China \and
\email{\{zhenyuli17, csxm, jiangjunjun\}@hit.edu.cn}\\
\email{lovesnow@mail.ustc.edu.cn 21B903029@stu.hit.edu.cn}}

\maketitle

\begin{abstract}
    Monocular depth estimation is an essential task in the computer vision community. While tremendous successful methods have obtained excellent results, most of them are computationally expensive and not applicable for real-time on-device inference. In this paper, we aim to address more practical applications of monocular depth estimation, where the solution should consider not only the precision but also the inference time on mobile devices. To this end, we first develop an end-to-end learning-based model with a tiny weight size (1.4MB) and a short inference time (27FPS on Raspberry Pi 4). Then, we propose a simple yet effective data augmentation strategy, called \textbf{R$^{2}$ crop}, to boost the model performance. Moreover, we observe that the simple lightweight model trained with only one single loss term will suffer from performance bottleneck. To alleviate this issue, we adopt multiple loss terms to provide sufficient constraints during the training stage. Furthermore, with a simple dynamic re-weight strategy, we can avoid the time-consuming hyper-parameter choice of loss terms. Finally, we adopt the structure-aware distillation to further improve the model performance. Notably, our solution named \textit{LiteDepth} ranks \textbf{2$^{nd}$ in the MAI\&AIM2022 Monocular Depth Estimation Challenge}, with a si-RMSE of 0.311, an RMSE of 3.79, and the inference time is 37$ms$ tested on the Raspberry Pi 4. Notably, we provide the \textbf{fastest} solution to the challenge. Codes and models will be released at \url{https://github.com/zhyever/LiteDepth}. 
 
\keywords{Monocular Depth Estimation, Lightweight Network, Data Augmentation, Multiple Loss}
\end{abstract}

\section{Introduction}

Monocular depth estimation plays a vital role in the computer vision community, where a wide spread of various depth-depended tasks related to autonomous driving~\cite{chen2022autoalign,wang2022monocular,chen2022graph,reading2021categorical,chen2022autoalignv2,weng2019monocular,li2022unsupervised,wang2022detr3d}, virtual reality~\cite{armbruster2008depth,gerig2018missing}, and scene understanding~\cite{zhu2020edge,wang2018depth,hazirbas2016fusenet,vu2019dada} provide strong demand for fast and accurate monocular depth estimation methods that are applicable to portable low-power hardware. Therefore, research along the line of accelerating depth estimation while reducing quality sacrifice on mobile devices has drawn increasing attention~\cite{ignatov2021fast,wang2021knowledge}.

As a classic ill-posed problem, estimating accurate depth from a single image is challenging. However, with the fast development of deep learning techniques, neural network demonstrates groundbreaking improvement with plausible depth estimation results~\cite{eigen2014depth,lee2019bts,bhat2021adabins,yang2021transdepth,li2022depthformer,li2022binsformer}. While engaging results have been presented, most of these state-of-the-art (SoTA) models are only optimized for high fidelity results while not taking into account computational efficiency and mobile-related constraints. The requirements of powerful high-end GPUs and consuming gigabytes of RAM lead to a dilemma when developing these models on resource-constrained mobile hardware~\cite{ignatov2021fast,huawei2018,snapdragon2018}.

In this paper, we aim to address the more practical application problem of monocular depth estimation on mobile devices, where the solution should consider not only the precision but also the inference time~\cite{ignatov2021fast}. We first investigate a suitable network design. Typically, the depth estimation network follows a UNet paradigm~\cite{ronneberger2015u} consisting of an encoder and a decoder with skip connections. Regarding the encoder, we choose a variant version of MobileNet-v3~\cite{howard2017mobilenets} as a trade-off between performance and inference time, where we drop out the last convolution layer to speed up inference and reduce the model size. Moreover, we observe that the commonly used image normalization pre-process on input images is also time-consuming (19$ms$ on Raspberry Pi 4). To solve this issue, we propose to merge the normalization into the first convolution layer in a post-process manner so that the redundant overhead can be eliminated without bells and whistles. Following~\cite{ignatov2021fast}, we adopt the fast downsampling strategy, which could quickly downsample the resolution of input images from 480 $\times$ 640 to 4 $\times$ 6. A light decoder is introduced to recover the spatial details, consisting of a few convolutional layers and upsampling layers. 

After determining the model structure, we propose several effective training strategies to boost the fidelity of the lightweight model. (1) We adopt an effective augmentation strategy called \textbf{R$^{2}$ crop}. It not only adopts crop patches on images with \textbf{R}andom locations but also \textbf{R}andomly changes the size of crop patches. This strategy increases the diversity of the scenes and effectively avoids overfitting the training set. (2) We introduce a multiple-loss training strategy to provide sufficient supervision during the training stage, where we propose a gradience loss that can handle invalid holes in training samples and adopt the other three loss terms proposed in previous works. Moreover, we install a dynamic re-weighting strategy that can avoid the time-consuming weight selection of loss terms. (3) We highlight that our work focuses on the model training strategies, unlike previous solutions that adopt variant distillation methods~\cite{ignatov2021fast,wang2021knowledge}. However, model distillation can also be an effective way to boost the model fidelity without any overhead. Therefore, we adopt the structure-aware distillation~\cite{liu2020structured} in a fine-tuning manner. 

We evaluate our method on Mobile AI (MAI2022) dataset, and the results demonstrate that each strategy can improve the accuracy of the lightweight network. With a short inference time (37$ms$ per image) on Raspberry Pi 4 and a lightweight model design (totally 1.4MB), our solution named \textit{MobileDepth} achieves results of 0.311 si-RMSE and ranks second in the MAI\&AIM 2022 Monocular Depth Estimation Challenge~\cite{ignatov2022depth}.

In summary, our main contributions are:
\begin{itemize}
    \item[$\bullet$] We design a lightweight depth estimation model that achieves fast inference on mobile hardware, where an image normalization merging strategy is proposed to reduce the redundant overhead.
    
    \item[$\bullet$] We adopt an effective augmentation strategy called R$^{2}$ crop that is adopted at random locations on images with a randomly changed size of patches.
    
    \item[$\bullet$] We design a gradience loss that can handle invalid holes in training samples and propose to apply multiple-loss items to provide sufficient supervision during the training stage.
    
    \item[$\bullet$] We evaluate our method on MAI2022 dataset and rank second place in the MAI\&AIM2022 Monocular Depth Estimation Challenge~\cite{ignatov2022depth}.
\end{itemize}

\begin{table*}[thbp]
    \caption{Ranking results in the MAI\&AIM2022 Monocular Depth Estimation Challenge, which are evaluated on the online test server. We highlight our results in \textbf{bold}.}
    \centering
        \begin{adjustbox}{width=0.95\linewidth,center}
            \begin{tabular}{@{}cccccccc@{}}
            \hline
            Rank & Username & ~si-RMSE~ & ~~~RMSE~~ & ~~~log$_{10}$~~~ & ~~~~REL~~~~ & ~Runtime~ & ~~Score~~ \\
            \hline
            1 & TCL & 0.277 & 3.47 & 0.110 & 0.299 & 46$ms$ & 297.79 \\ 
            \textbf{2} & \textbf{Zhenyu Li} & \textbf{0.311} & \textbf{3.79} & \textbf{0.124} & \textbf{0.342} & \textbf{37$ms$} & \textbf{232.04} \\
            3 & ChaoMI & 0.299 & 3.89 & 0.134 & 0.380 & 54$ms$ & 187.77 \\
            4 & parkzyzhang & 0.303 & 3.80 & 12.189 & 0.301 & 68$ms$ & 141.07 \\
            5 & RocheL & 0.329 & 4.06 & 0.137 & 0.366 & 65$ms$ & 102.07 \\
            6 & mvc & 0.349 & 4.46 & 0.140 & 0.340 & 139$ms$ & 36.07 \\ 
            7 & Byung Hyun Lee & 0.338 & 6.73 & 0.332 & 0.507 & 142$ms$ & 41.58 \\
            \hline
            \end{tabular}
        \end{adjustbox}
    \label{tab::overall}
\end{table*}

\section{Related Work}

Monocular depth estimation is an ill-posed problem~\cite{eigen2014depth}. Lack of cues, scale ambiguities, translucent or reflective materials all leads to ambiguous cases where appearance cannot infer the spatial construction~\cite{li2022depthformer}. With the rapid development of deep learning, the neural network has dominated the primary workhorse to provide reasonable depth maps from a single RGB input~\cite{yin2019enforcing,lee2019bts,bhat2021adabins,li2022depthformer,li2022binsformer}. 

Eigen et al.~\cite{eigen2014depth} first groundbreakingly propose a multi-scale deep network, consisting of a global network and a local network to predict the coarse depth and refine predictions, respectively. Subsequent works focus on various points to boost depth estimation, for instance, problem formulation~\cite{fu2018deep,bhat2021adabins,li2022binsformer}, network architecture~\cite{lee2019bts,li2022depthformer,kim2022global}, supervision design~\cite{barron2019general,yin2019enforcing,patil2022p3depth}, interpretable method~\cite{you2021towards}, pre-training strategy~\cite{li2021simipu,park2021pseudo}, unsupervised training~\cite{godard2019digging,zhou2017unsupervised}, \textit{etc}. Though achieving engaging fidelity, these methods neglect the limitation of resource-constrained hardware and can be hard to develop on portable devices or embedded systems. 

Notably, there are also some methods that take the inference time and model complexity into account, which makes them applicable on mobile devices~\cite{ignatov2021fast}. FastDepth~\cite{wofk2019fastdepth} deploys a real-time depth estimation method on embedded systems by designing an efficient model architecture and a pruning strategy to further reduce the model complexity. In our paper, we follow FastDepth~\cite{wofk2019fastdepth} to choose MobileNet-v3~\cite{howard2017mobilenets} as our encoder and design an even more lightweight decoder (only consisting of four convolution layers) to achieve a trade-off between fidelity and inference speed.

\section{Method}
In this section, we first present our network design in Sec.~\ref{sec::network_design}, where tons of details should be considered to achieve the best trade-off between fidelity and inference speed. Then, we introduce our proposed R$^2$ Crop in Sec.~\ref{sec::r2_crop} and Multiple Loss Training strategy in Sec.~\ref{sec::multi_loss}. Subsequently, we illustrate the installation of the structure-aware Distillation strategy in Sec.~\ref{sec::distill}.

\begin{figure*}[t]
    \centering
      \includegraphics[width=1\linewidth]{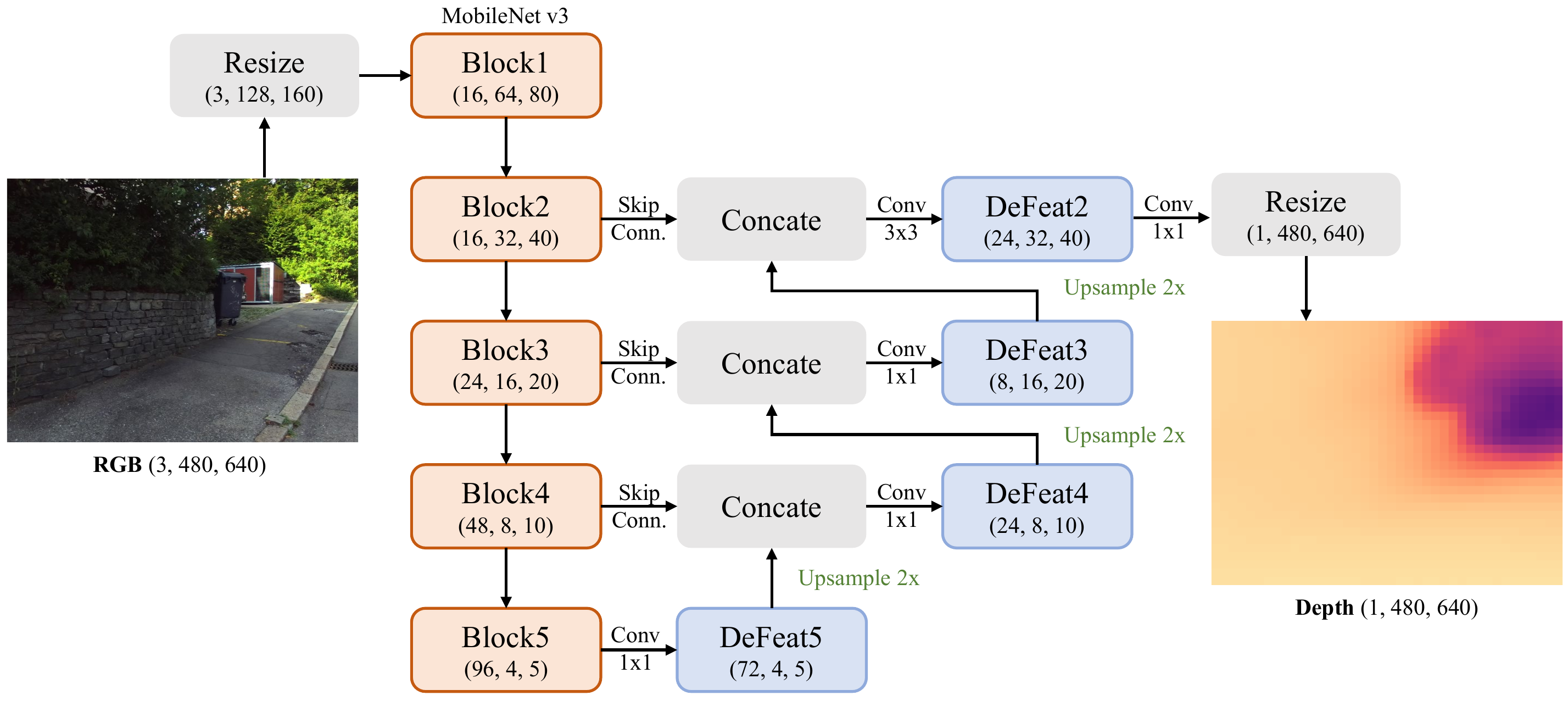}
      \caption{Illustration of our proposed network architecture that follows the prevalent Unet~\cite{ronneberger2015u} design consisting of a MobileNet-V3~\cite{howard2017mobilenets} encoder and a lightweight decoder with skip connections.}
    \label{fig::overall}
\end{figure*} 
  
\subsection{Network Design}
\label{sec::network_design}

As shown in Fig.~\ref{fig::overall}, our proposed network consists of an encoder and a lightweight decoder with skip connections. We sequentially introduce each component and design detail.

\textbf{Encoder} plays a crucial role in extracting features from input images for depth estimation. To achieve a trade-off between fidelity and inference speed, we choose MobileNet-v3~\cite{howard2017mobilenets} as our encoder. It is worth noticing that MobileNet contains a dimension-increasing layer (1$\times$1 convolution with an input dimension of 96 and output dimension of 960) to facilitate training for a classification task. We remove this layer to improve the inference speed and reduce the number of model parameters. Following~\cite{ignatov2021fast}, we adopt the \textit{Fast Downsampling Strategy} in which a resize layer is inserted at the beginning of the encoder to resize the high-resolution input image from 480 × 640 to 128 × 160. As a result, the encoder can quickly downsample the resolution of feature maps, significantly shorten the inference time. Typically, input images are normalized to align with the pre-training setting. We discern the vanilla image normalization is time-consuming (19$ms$ of the image normalization \textit{v.s.} 37$ms$ of the whole model) on the target device (\textit{i.e.,} Raspberry Pi 4). Therefore, we propose to merge the image normalization into the first convolution layer in a post-process manner so that we can avoid the redundant overhead \textit{without bells and whistles}. Consider the image normalization and the first convolution layer:

\begin{equation}
  I_n = \frac{I_r - m}{s},
\end{equation}
\begin{equation}
  f = W * I_n + b,
\end{equation}
where $I_n$ and $I_r\in\mathbb{R}^{3\times H\times W}$ are normalized and raw input images. $m\in\mathbb{R}^{3}$ and $s\in\mathbb{R}^{3}$ are the mean and standard deviation used in the image normalization. $f\in\mathbb{R}^{C\times H_f \times W_f}$ is the output feature map with $C$ channels of the first convolution in our network. $W\in\mathbb{R}^{3\times C\times k^2}$ and $b\in\mathbb{R}^{C}$ are the trained weight and bias of the first $k\times k$ convolution. $*$ denotes the convolution operation. Given a trained model with parameters $W$ and $b$ of the first convolution, we update the them based on the mean and standard deviation used in the image normalization during the training stage:

\begin{equation}
  W' = \frac{W}{s},
\end{equation}
\begin{equation}
  b_i' = b_i - \sum\limits_{d}^{3}\left(\frac{m_d}{s_d}\times \sum\limits_{j}^{k\times k}W_{dij}\right),~i\in(1, 2, ..., C),
\end{equation}
\begin{equation}
  b' = \mathbf{Concat}([b_1', b_2', ..., b_C']),
\end{equation}
where the $W'$ and $b'$ are the updated weight and bias of the first $k\times k$ convolution. $\mathbf{Concat}$ is the element-wise concatenation. $d$ is the index of RGB dimension. Consequently, we discard the image normalization and apply the first convolution directly on input images as:
\begin{equation}
  f = W' * I_r + b'.
\end{equation}
As a result, the trained network can directly recieve the raw input images without the time-consuming image normalization.

\textbf{Decoder} is adopted to recover the spatial details by fusing the multi-level deep and shallow features. Unlike previous works~\cite{wofk2019fastdepth,ignatov2021fast,wang2021knowledge} that utilize the symmetrical encoder and decoder, we drop out the last decoder layer to further accelerate the model inference. Hence, the resolution of outputs is 4$\times$ downsampled (\textit{i.e.,} 32$\times$64).  At each decoding stage, we apply a simple feature fusion module to aggregate the decoded and skip-connected features, which consists of a concatenation operation and a convolution layer (with ReLU as the activation function). To achieve the best trade-off between fidelity and speed, we utilize the 1$\times$1 and 3$\times$3 convolution for deep and shallow features, respectively. The final feature map is projected to the predicted depth map via the 1$\times$1 convolution, which is then passed by a ReLU function to suppress the plural prediction. Finally, we insert a resize block at the end of the decoder to upsample the predicted depth map to the raw resolution 480$\times$640. We highlight the lightweight design of the decoder that \textit{only consists of five convolution layers} but achieves satisfactory fidelity.

\begin{figure*}[t!]
  \centering
  \footnotesize
  \begin{tabular}{@{}ccc@{}}
      \includegraphics[width=0.25\linewidth]{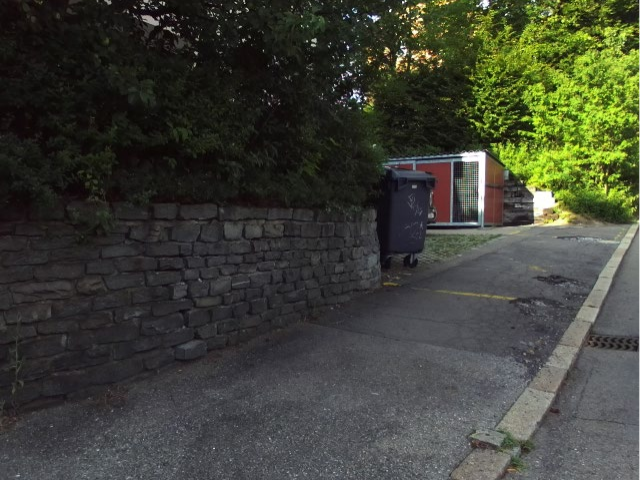} & 
      \includegraphics[width=0.25\linewidth]{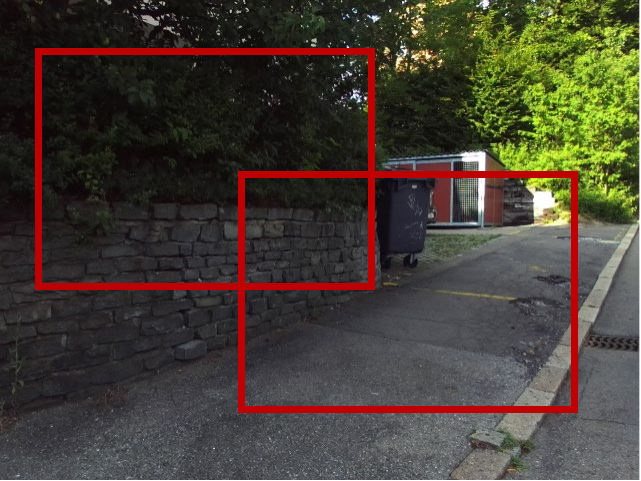} & 
      \includegraphics[width=0.25\linewidth]{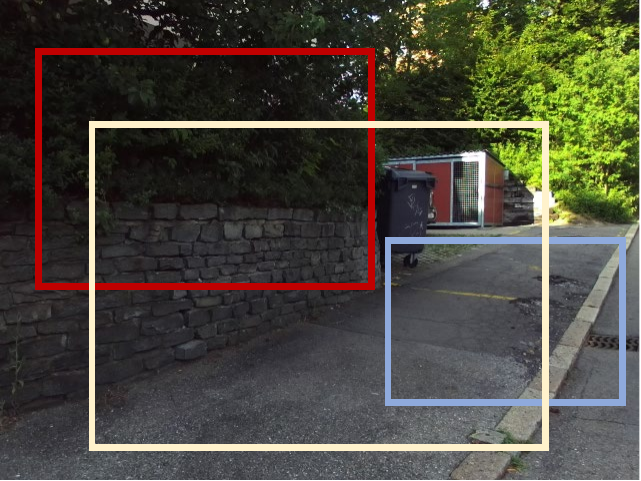} \\
      w/o crop & random crop & R$^2$ crop \\
  \end{tabular}
  \caption{Comparisons among different crop augmentations. As for R$^2$ crop, we utilize different colors to indicate that we adopt randomly selected size of crop patches.}
  \label{fig::crop_compare}
\end{figure*}

\begin{figure*}[t]
  \centering
    \includegraphics[width=1\linewidth]{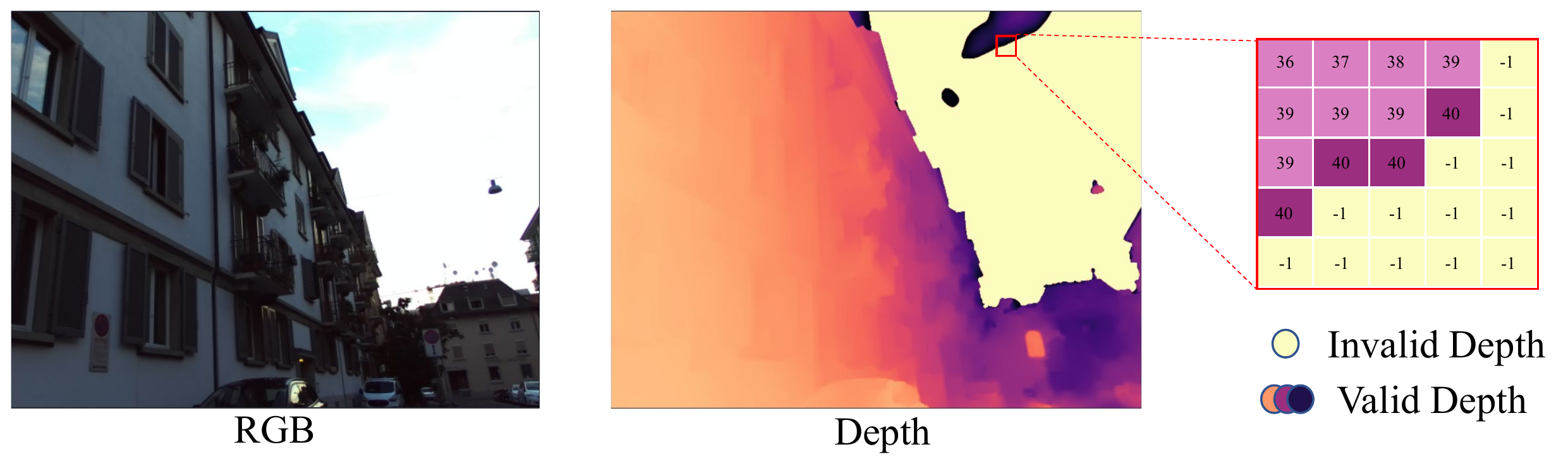}
    \caption{Illustration of invalid depth GT pixels in the dataset. These pixels appear not only in the sky areas but also in close positions where the sensor cannot provide reliable GT value. We highlight a training sample for a clear Introduction of our valid mask in gradloss in Fig.~\ref{fig::gradloss_2}.}
  \label{fig::gradloss_1}
\end{figure*} 

\subsection{R$^2$ Crop}
\label{sec::r2_crop}

Data augmentation is crucial to training models with better performance. Typically, the sequence of data augmentation for monocular depth estimation includes random rotation, random flip, random crop, and random color enhancement~\cite{lidepthtoolbox2022}. We propose the more effective crop strategy R$^2$ crop, in which we randomly select the size of crop patches and the cropped locations. We highlight the discrepancy with other commonly used crop methods in Fig.~\ref{fig::crop_compare}. It increases the diversity of the scenes and effectively avoids overfitting the training set.

\subsection{Multiple Loss Training}
\label{sec::multi_loss}

Previous depth estimation methods~\cite{lee2019bts,bhat2021adabins,li2022depthformer,li2022binsformer} only adopt the silog loss to train the neural network:
\begin{equation}
  \mathcal{L}_{silog} = \alpha \sqrt{\frac{1}{N}\sum\limits_i^N e_{i}^{2} - \frac{\lambda}{N^2}(\sum\limits_i^N e_i)^2},
  \label{eq:silog_loss}
\end{equation}
where $e_i = \log \hat{d}_i - \log d_i$ with the ground truth depth $d_i$ and predicted depth $\hat{d}_i$. $N$ denotes the number of pixels having valid ground truth values. Since we discover that the lightweight model supervised by this simple single loss lacks representation capability and is easily stuck in local optimal, we adopt diverse loss terms to provide various targets for sufficient model training.

\begin{figure*}[t]
  \centering
    \includegraphics[width=1\linewidth]{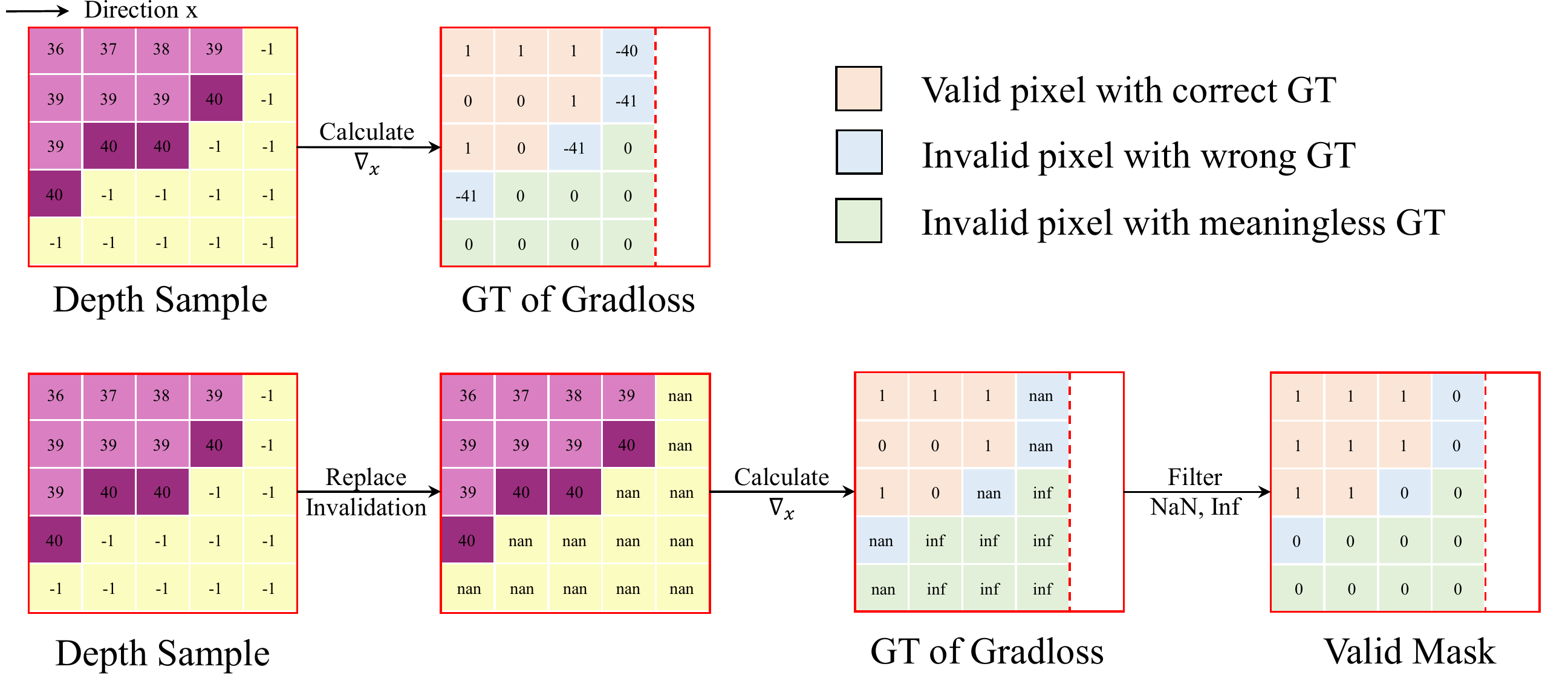}
    \caption{Illustration of valid mask calculation for gradience loss (x direction). First line: vanilla calculation of gradience loss. Second line: we propose to first replace invalid value with NaN and compute reasonable valid mask for gradience loss.}
  \label{fig::gradloss_2}
\end{figure*}

Motivated by~\cite{sitzmann2020implicit}, we first propose a \textbf{gradience loss} $\mathcal{L}_{grad}$ formulated as:

\begin{equation}
  \mathcal{L}_{grad} = \frac{1}{N}\sum_i\left(M_{x_i}\times \left\|\nabla_x\hat{d}_i - \nabla_x d_i\right\|_1 + M_{y_i}\times \left\|\nabla_y\hat{d}_i - \nabla_y d_i\right\|_1\right),
  \label{eq:grad_loss}
\end{equation}
where $\nabla$ is the gradience calculation operation. Since the gradience loss is calculated in a dislocation subtraction manner and there are tremendous invalid depth GT in the dataset as shown in Fig.~\ref{fig::gradloss_1}, as presented in Fig.~\ref{fig::gradloss_2}, simply applying gradience calculation will blemish the information of invalid pixels and introduce outlier values when calculating the loss term. Hence, it is necessary to carefully design a strategy to calculate masks $M$ to filter these invalid pixels in $\mathcal{L}_{grad}$. To solve this issue, we first replace the invalid value with \textit{NaN} and then calculate the GT for gradience loss. Thanks to the numeral property of \textit{NaN} and \textit{Inf}, invalid information can be reserved. Consequently, we can filter the \textit{NaN} and \textit{Inf} when calculating the gradience loss.

Moreover, we also adopt the \textbf{virtual norm loss} $\mathcal{L}_{vnl}$~\cite{yin2019enforcing}, and \textbf{robust loss} $\mathcal{L}_{robust}$~\cite{barron2019general}. We formulate them as follows:

\begin{equation}
  \mathcal{L}_{vnl} = \frac{1}{N}\sum\limits_i^N\left\|\hat{n}_i - n_i \right\|_1,
  \label{eq:vnl_loss}
\end{equation}
where $n$ is the virtual norm. We refer more details in the original paper~\cite{yin2019enforcing}. Unlike the original implementation, we sample points from reconstructed point clouds and adopt constraints on predictions to filter invalid samples instead of ground truth. It helps the model convergence at the beginning of training.

\begin{equation}
  \mathcal{L}_{robust} = \frac{1}{N}\sum\limits_i^N\frac{|\alpha - 2 |}{\alpha}\left(\left(\frac{(e_i/c)^2}{|\alpha - 2|} \right)^{\alpha/2}-1 \right),
  \label{eq:robust_loss}
\end{equation}
where $e_i = \hat{d}_i - d_i$. We experimentally set $\alpha=1$ and $c=2$. In fact, the loss reduces to a simple $L_2$ loss, but which is proven to be more effective compared with the proposed adaptive version in our task. More experiments can be conducted to decide a better choice for $\alpha$ and $c$.

Finally, we adopt a combination of these loss terms to train our network. The total depth loss is

\begin{equation}
  \mathcal{L}_{depth} = w_1\mathcal{L}_{silog} + w_2\mathcal{L}_{grad} + w_3\mathcal{L}_{vnl} + w_4\mathcal{L}_{robust}.
\end{equation}

We set $w_1=1$, $w_2=0.25$, $w_3=2.5$, and $w_4=0.6$ based on tremendous experiments. Then, we apply a dynamic re-weight strategy in which the loss weights $w$ are set as model parameters and are automatically fine-tuned during the training stage. Experimental results indicate that this strategy can achieve similar results as tuning the weights by hand.

\begin{figure*}[t]
  \centering
    \includegraphics[width=1\linewidth]{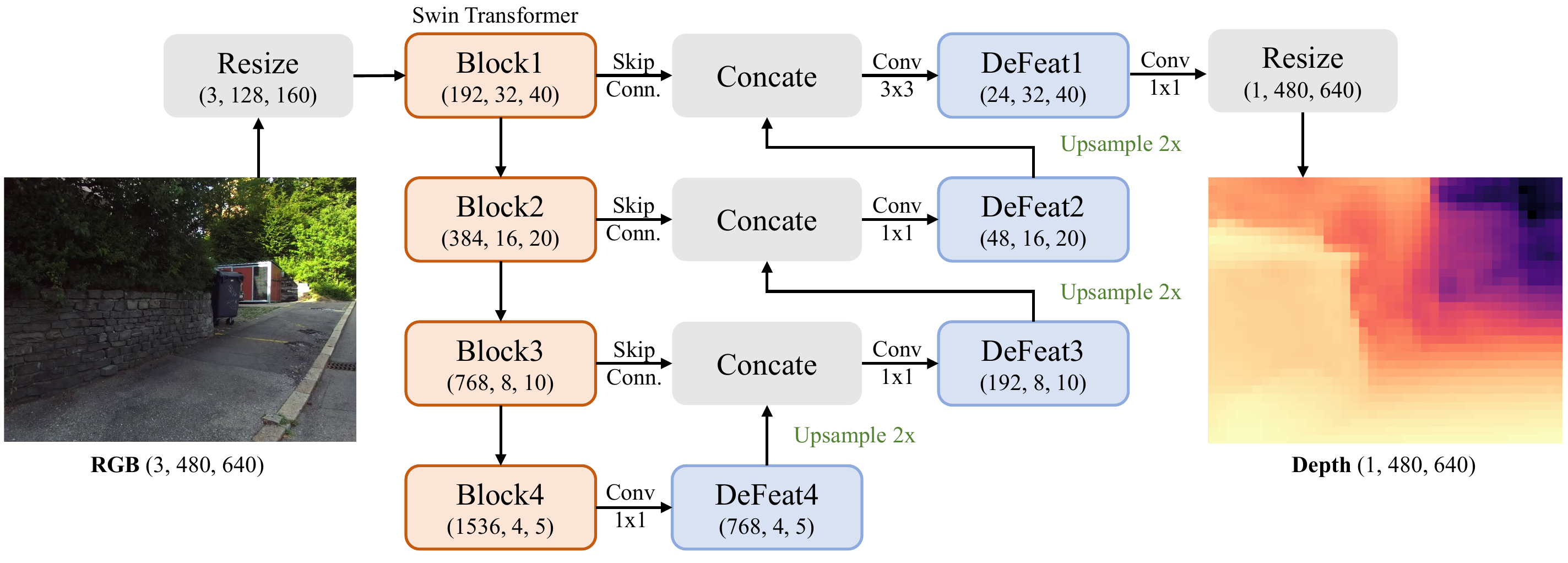}
    \caption{Illustration of the teacher network.}
  \label{fig::network_teacher}
\end{figure*} 

\subsection{Structure-Aware Distillation}
\label{sec::distill}

\begin{figure*}[t]
  \centering
    \includegraphics[width=1\linewidth]{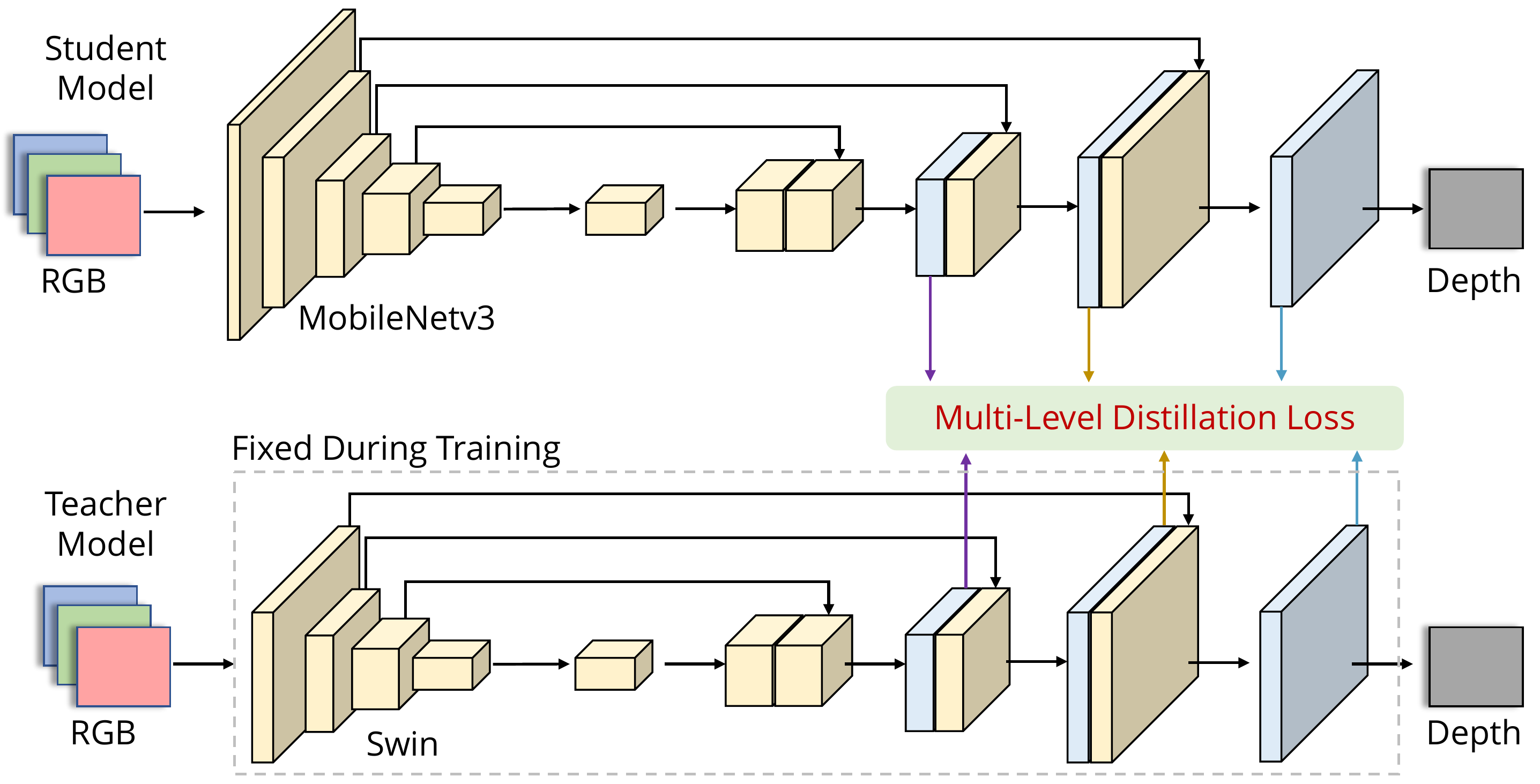}
    \caption{Illustration of our multi-scale distillation strategy.}
  \label{fig::distill}
\end{figure*}

We apply the structure-aware distillation strategy~\cite{liu2020structured,wang2021knowledge} to further boost model performance. For the teacher model, we choose Swin Transformer~\cite{liu2021swin} as the encoder and adopt a similar lightweight decoder to recover featrue resolution and predict depth maps. We present the network architecture in Fig.~\ref{fig::network_teacher}. The teacher model is trained via the supervision of $\mathcal{L}_{depth}$ and is then fixed when distilling the student model. During the distillation, multi-level distlling losses are adopted to provide supervisons on immediate features as shown in Fig.~\ref{fig::distill}. The distillation loss is formulated as
\begin{equation}
  \mathcal{L}_{distill} = \sum\limits_l^L\left(\frac{1}{H\times W}\sum\limits_i^H\sum\limits_j^W\left\|a_{ij}^s - a_{ij}^t \right\|_1\right),
\end{equation}
where $a$ is the affinity map calculated via inner-product of $L_2$ normalized features. We refer to \cite{liu2020structured,wang2021knowledge} for more details. $s$ and $t$ indicate the features are from the student and teacher model, respectively. We choose three level ($L=3$) features for distll, which are DeFeat2, DeFeat3, and DeFeat4 in Fig.~\ref{fig::overall}. 

Consequently, the student model is trained via the total loss $\mathcal{L}$:
\begin{equation}
  \mathcal{L} = \mathcal{L}_{depth} + w_d\mathcal{L}_{distill},
\end{equation}
where $w_d=10$ in our experiments. Notably, unlike previous work~\cite{liu2020structured,wang2021knowledge}, we adopt a two-stage training paradigm. During the first stage, the student model is only trained via $\mathcal{L}_{depth}$. In the second stage, we adopt the teacher model and utilize the total loss $\mathcal{L}$ to further boost the performance of the student model.

\section{Experiments}

In this section, we introduce our experiments to evaluate the effectiveness of our solution. We first elaborate the dataset and define the evaluation metrics. Then the detailed implementation and ablation studies are presented. We also report the inference time on target devices (\textit{i.e.}, Raspberry Pi 4) to show that our method can not only produce reasonable depth estimation but also achieve real-time inference on resource-constrained hardware.

\subsection{Setup}

\subsubsection{Dataset}

We utilize the dataset provided by MAI\&AIM2022 challenge to conduct experiments, which contains 7385 pairs of RGB and grayscale depth images. The pixel values of depth maps are in uint16 format ranging from 0 to 40000, which represent depth values from 0 to 40 meters. We use 6869 pairs for training and the rest 516 pairs as the local validation set. 

\subsubsection{Evaluation Metrics}

In MAI\&AIM2022 challenge~\cite{ignatov2022depth}, two metrics are considered for each submission solution: 1) The quality of the depth estimation. It is measured by the invariant standard root mean squared error (si-RMSE). 2) The runtime of the model on the target platform (\textit{i.e.,} Raspberry Pi 4). The scoring formulation is provided below:

\begin{equation}
    \text{Score}(\text{si-RMSE}, \text{runtime}) = \frac{2^{-20}\cdot \text{si-RMSE}}{C\cdot \text{runtime}},
\end{equation}
where $C=0.01$ on the online validation benchmark.

\subsection{Implementation Details}

We implement the proposed model via the monocular depth estimation toolbox~\cite{lidepthtoolbox2022}, which is based on the open-source machine learning library Pytorch. The model is converted to TFLite~\cite{lite2019deploy} after training. We use Adam optimizer with betas = (0.9, 0.999) and eps=1e-3. A poly schedule is adopted where the base learning rate is 4e$^{-3}$ and the power is 0.9. The total number of epochs is 600 with batch size = 32 on two RTX3090 GPUs, which takes around 4 hours to train a model. The encoder of our network is pretrained on ImageNet, and the decoder part is trained from scratch.

\begin{figure*}[t]
  \centering
    \includegraphics[width=0.8\linewidth]{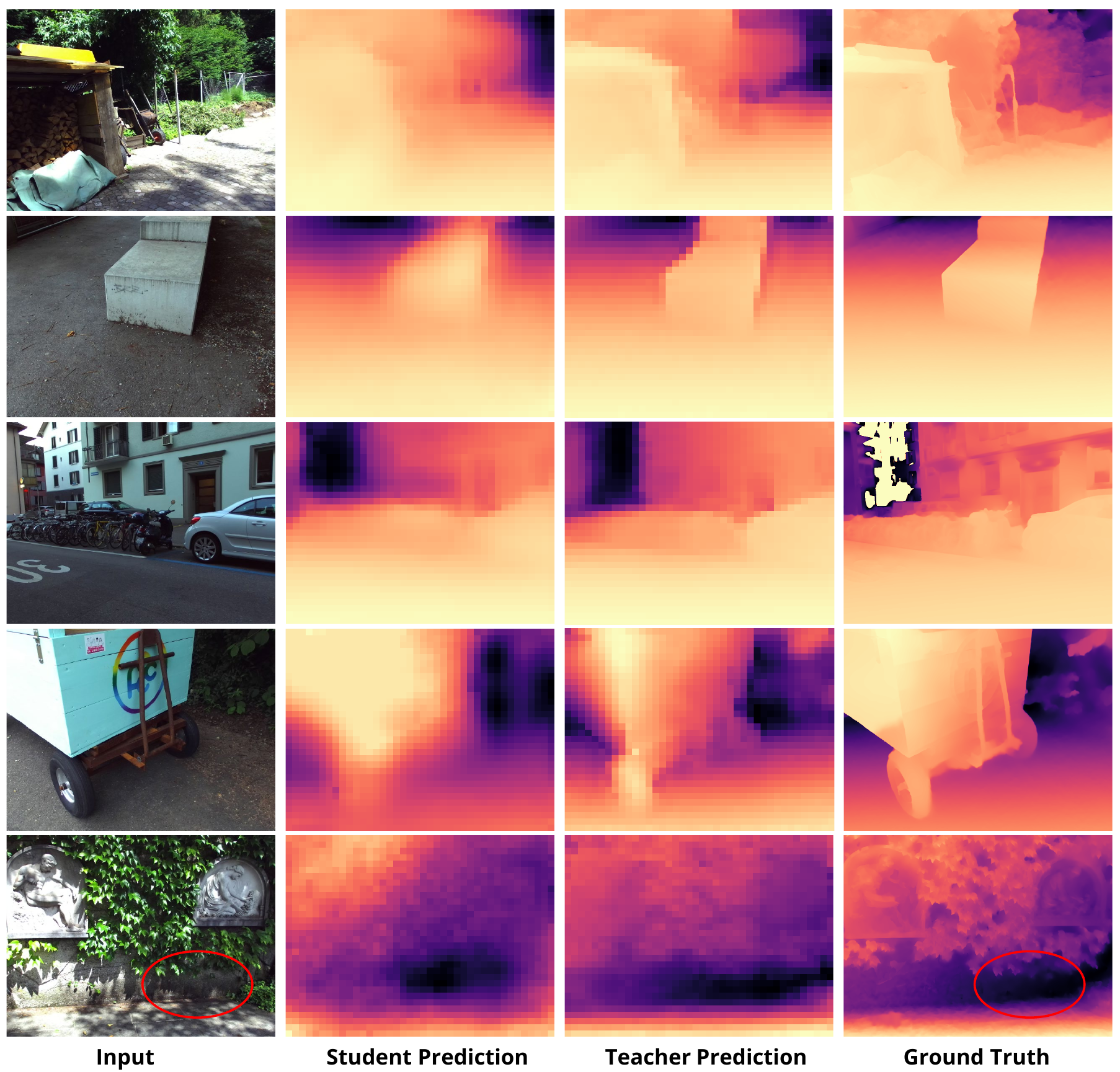}
    \caption{The visualization results of our proposed methods. One can observe that there is noise in ground truth labels which we highlight with a red circle.}
  \label{fig::val_figures}
\end{figure*}

\subsection{Quantitative Results}

As shown in Tab.~\ref{tab::overall}, our proposed method achieves a score of 232.04 on the challenge test set and ranks second place. Our solution achieves 0.311 si-RMSE with 37$ms$ on the Raspberry Pi 4. Notably, our runtime is lower than the other methods and the performance is comparable.

\subsection{Qualitative Results}

We visualize the prediction results of our proposed methods as shown in Fig.~\ref{fig::val_figures}, which demonstrates that our methods can achieve reasonable depth estimation results. However, the predicted depth maps are very rough around the edges due to the excessive down-sampling.

\subsection{Inference Time}
In this section, we verify that our method can achieve high-throughput monocular depth estimation on mobile devices. We convert our model to TensorFlow-Lite and test the inference time on various mobile devices, including smartphones with Kirin 980 and Snapdragon 7 Gen 1. We test the model using AI Benchmark~\cite{ignatov2018ai,ignatov2019ai}. Following the challenge requirements, the resolution of input and output images is 640$\times$480. The data type is set to float (32 bit). As presented in Tab~\ref{tab::runtime}, our network can obtain extremely high-throughput inference. It achieve 162FPS on smart phones with Snapdragon 7 Gen 1 processor. Interestingly, we can observe that the model is CPU-friendly, with an even faster inference on CPU than GPU on mobile devices.

\begin{table*}[thbp]
    \caption{Inference time of our network (AI Benchmark).}
    \centering
        \begin{adjustbox}{width=0.80\linewidth,center}
            \begin{tabular}{@{}llcc@{}}
            \hline
            SoC  & Device & Average/ms & ~~STD/ms~~ \\
            \hline
            Kirin 980 & CPU & 6.85 & 0.77 \\
            Kirin 980 & GPU Delegate & 9.84 & 0.66 \\
            \hline
            Snapdragon 7 Gen 1~~ & CPU & 6.16 & 1.71 \\
            Snapdragon 7 Gen 1~~ & GPU Delegate~~ & 7.17 & 1.00 \\
            \hline
            \end{tabular}
        \end{adjustbox}
    \label{tab::runtime}
\end{table*}

\subsection{Ablation Studies}


\subsubsection{Effectiveness of Network Design}

Encoder selection is crucial to the trade-off between fidelity and runtime. We recommend refering~\cite{ignatov2021fast,wang2021knowledge} for more comparisons among various encoders. Following these works, we choose the MobileNet-v3 as the default encoder. We then present comparisons among different decoder designs as shown in Tab.~\ref{tab::network_arch}. Typically, previous methods~\cite{ignatov2021fast,wang2021knowledge,lee2019bts,bhat2021adabins,li2022depthformer} utilize the 3$\times$3 convolution to fuse features. While the quantitative results are good, the runtime can be longer. However, when we replace all the 3$\times$3 convolution with 1$\times$1 convolution, the model performance drops drastically while the runtime gets short. Hence, we adopt a \textbf{\textit{mix}} version as presented in our Sec.~\ref{sec::network_design} and Fig.~\ref{fig::overall}. We utilize a 3$\times$3 convolution at the highest resolution and adopt 1$\times$1 convolutions at other places, which makes the best trade-off between fidelity and runtime, getting the highest score on the benchmark. We then present the importance of the \textit{merging image normalization}. It significantly reduces the runtime without any performance drop.

\begin{table*}[htbp]
    \caption{Ablation study about the network architecture design. Dec and MIN are the short for decoder and \textit{merge image normalization}, respectively.}
    \centering
        \begin{adjustbox}{width=0.8\linewidth,center}
            \begin{tabular}{@{}lc|ccc@{}}
            \hline
            Architecture & ~~~MIN~~~ & ~~si-RMSE~~ & Runtime/ms & ~~~Score~~~ \\
            \hline
            Full 3$\times$3 @ Dec. && 0.295 & 62 & 27.01\\
            Full 1$\times$1 @ Dec. && 0.308 & 53 & 26.38\\
            Mix Convs @ Dec. && \textbf{0.301} & 56 & 27.51\\
            Mix Convs @ Dec. &\ding{51} & \textbf{0.301} & \textbf{37} & \textbf{41.64}\\
            \hline
            \end{tabular}
        \end{adjustbox}
    \label{tab::network_arch}
  \end{table*} 

\subsubsection{Effectiveness of R$^2$ Crop}

We present the ablation study of various crop strategies. In these experiments, we only adopt the single sigloss (Eq.~\ref{eq:silog_loss}) for simplicity. As shown in Tab.~\ref{tab::ablation_crop}, our proposed R$^2$ crop indicates an engaging improvement on performance compared with the baseline methods. When we adopt the vanilla random crop, the model cannot learn the knowledge of full-area images. However, the model infers on full-area images during the validation stage. This discrepancy leads to significant performance degradation. If we do not apply any crop strategy, the diversity of training samples is limited, also leading to a performance limitation. When we adopt our proposed R$^2$ crop, during the training stage, the model can not only learn the knowledge of full-area images but also ensure the diversity of training samples. When increasing the variety of crop sizes, the model performance can be improved simultaneously. However, too small patches cannot bring performance gains but lead to a slight degradation (\textit{e.g.,} (144, 256) patches in our ablation study). We infer that the small patches do not contain sufficient structure information for facilitating the model training. As a result, we adopt patches with a size of [(240, 384), (384, 512), (480, 640)] in our solution.

\begin{table*}[thbp]
    \caption{Ablation study of crop strategies. (h, w) represents the size of crop patches.}
    \centering
        \begin{adjustbox}{width=0.9\linewidth,center}
            \begin{tabular}{@{}l|cc@{}}
            \hline
            Method  & ~~si-RMSE~~ & RMSE \\
            \hline
            w/o crop & 0.335 & 4.25 \\
            random crop with (384, 512) & 0.377 & 4.62 \\
            R$^2$ crop with [(384, 512), (480, 640)] & 0.327 & 4.15 \\
            R$^2$ crop with [(240, 384), (384, 512), (480, 640)] & \textbf{0.323} & \textbf{4.11} \\
            R$^2$ crop with [(144, 256), (240, 384), (384, 512), (480, 640)] & 0.325 & 4.13 \\
            \hline
            \end{tabular}
        \end{adjustbox}
    \label{tab::ablation_crop}
  \end{table*}

\subsubsection{Effectiveness of Multiple-Loss Training}

This section evaluates the effectiveness of each loss term used in our solution. The results are presented in Tab.~\ref{tab::ablation_multi_loss}. Each loss term can bring performance gains for the model. We also highlight that if we do not apply the invalid mask in gradience loss, the model convergence will be hurt as described in Sec.~\ref{sec::multi_loss}. Moreover, our dynamic weight strategy can also achieve satisfactory results without fine-tuning loss weights by hand. We utilize the handcrafted weights as a default setting to achieve a better score in the challenge.

\begin{table*}[thbp]
    \caption{Ablation study of the multiple loss strategy.}
    \centering
        \begin{adjustbox}{width=0.95\linewidth,center}
            \begin{tabular}{@{}ccccc|cc@{}}
            \hline
            Sig Loss (Eq.~\ref{eq:silog_loss})& Grad Loss (Eq.~\ref{eq:grad_loss})& VNL Loss (Eq.~\ref{eq:vnl_loss}) & Robust Loss (Eq.~\ref{eq:robust_loss}) & Dynamic Weight & ~~si-RMSE~~  \\
            \hline
            \ding{51} & & & & & 0.323 \\
            \ding{51} & \ding{51} & & & & 0.316 \\
            \ding{51} & \ding{51} & \ding{51} & & & 0.309 \\
            \ding{51} & \ding{51} & \ding{51} & \ding{51} & & \textbf{0.303} \\
            \ding{51} & \ding{51} & \ding{51} & \ding{51} & \ding{51} & 0.306 \\
            \hline
            \end{tabular}
        \end{adjustbox}
    \label{tab::ablation_multi_loss}
  \end{table*}

\subsubsection{Effectiveness of Distillation}

We first present the results of the teacher model. As shown in Tab.~\ref{tab::ablation_distll}, the teacher model achieves much better fidelity compared to the student model. It indicates that there is improvement room for the student model to learn from the teacher model via the distillation. We also present qualitative results in Fig.~\ref{fig::val_figures} for intuitive comparisons. As we can observe from the predicted depth maps, the teacher model provides more reasonable and sharper depth estimation results.

We then evaluate different distillation strategies in this section. Motivated by previous work, we try to apply L2 distillation~\cite{ignatov2021fast}, structure-aware distillation~\cite{liu2020structured,wang2021knowledge}, and channel-wise distillation~\cite{shu2021channel}. Interestingly, all strategies cannot directly work well for our lightweight student model as presented in Tab.~\ref{tab::ablation_distll}. One possible reason is that we adopt multiple loss terms, leading to difficulty in balancing the loss weights. However, we also conduct experiments in which we only adopt the single sigloss and apply the distillation strategies. The results are similar without improvement in model performance. Moreover, some distillation strategies conflict with the two-stage fine-tuning, leading to a convergence issue. These experimental results indicate that more effective distillation strategies should be designed for monocular depth estimation. In this solution, we propose to adopt structure-aware distillation. It brings a slight improvement to the si-RMSE of our lightweight student model but a degradation on RMSE, indicating there is still huge room to improve the distillation strategy.

\begin{table*}[thbp]
    \caption{Ablation study of distillation strategies. Two-Stage indicates applying the distillation in a fine-tuning manner. $\varnothing$ denotes that the fine-tuning process does not converge.}
    \centering
        \begin{adjustbox}{width=0.7\linewidth,center}
            \begin{tabular}{@{}lc|ccc@{}}
            \hline
            Method  & ~~Two-Stage~~ & ~~si-RMSE~~ & RMSE  \\
            \hline
            Teacher Model & & 0.228 & 3.025\\
            Baseline Student Model &  &  0.303 & \textbf{3.785} \\
            \hline
            L2 Distillation & & 0.307 & 3.978 \\ 
            L2 Distillation &\ding{51} & \multicolumn{2}{c}{~~~~$\varnothing$} \\
            Channel-Wise Distillation & & 0.311 & 4.045 \\
            Channel-Wise Distillation & \ding{51} & \multicolumn{2}{c}{~~~~$\varnothing$}\\
            Structure-Aware Distillation &  & 0.306 & 3.994 \\
            Structure-Aware Distillation & \ding{51} & \textbf{0.301} & 3.839 \\
            \hline
            \end{tabular}
        \end{adjustbox}
    \label{tab::ablation_distll}
  \end{table*}

\section{Conclusion}

We have introduced our solution for fast and accurate depth estimation on mobile devices. Specifically, we design an extremely lightweight model for depth estimation. Then, we propose R$^2$ crop to enrich the diversity of training samples. To facilitate the model training, we design a gradience loss and adopt multiple-loss items. We also investigate various distillation strategies. Extensive experiments indicate the effectiveness of our proposed solution.

\section{Acknowledgments}
The research was supported by the National Natural Science Foundation of China (61971165, 61922027), and also is supported by the Fundamental Research Funds for the Central Universities.


\bibliographystyle{splncs04}
\bibliography{ref}

\end{document}